\documentclass[conference]{IEEEtran}
\IEEEoverridecommandlockouts

\usepackage{cite}
\usepackage{amsmath,amssymb,amsfonts}
\usepackage{algorithmic}
\usepackage{graphicx}
\usepackage{textcomp}
\usepackage{xcolor}
\def\BibTeX{{\rm B\kern-.05em{\sc i\kern-.025em b}\kern-.08em
    T\kern-.1667em\lower.7ex\hbox{E}\kern-.125emX}}
\begin{document}

%
%
%
\title{Bayesian Active Inference for Intelligent UAV Anti-Jamming and Adaptive Trajectory Planning}

%
%
\author{Ali Krayani\textsuperscript{1,2}, Seyedeh Fatemeh Sadati\textsuperscript{1}, Lucio Marcenaro\textsuperscript{1,2}, and Carlo Regazzoni\textsuperscript{1,2}
\IEEEauthorblockA{\\ \textsuperscript{1}University of Genoa, DITEN, Italy\\
\textsuperscript{2}Italian National Inter-University Consortium for Telecommunications (CNIT), Italy
\\ \small \textit{ali.krayani@ieee.org}, \textit{sadatisogand@gmail.com}, \textit{lucio.marcenaro@unige.it}, \textit{carlo.regazzoni@unige.it}
}
}


\maketitle

\begin{abstract}
This paper proposes a hierarchical trajectory planning framework for UAVs operating under adversarial jamming conditions. Leveraging Bayesian Active Inference, the approach combines expert-generated demonstrations with probabilistic generative modeling to encode high-level symbolic planning, low-level motion policies, and wireless signal feedback. During deployment, the UAV performs online inference to anticipate interference, localize jammers, and adapt its trajectory accordingly—without prior knowledge of jammer locations. Simulation results demonstrate that the proposed method achieves near-expert performance, significantly reducing communication interference and mission cost compared to model-free reinforcement learning baselines, while maintaining robust generalization in dynamic environments.
\end{abstract}


\section{Introduction}
Unmanned Aerial Vehicles (UAVs) play a crucial role in military, public, and civilian applications due to their compact size, flexible deployment capabilities, and outstanding performance. These vehicles are utilized for various purposes, including low-altitude surveillance, post-disaster rescue operations, logistics support, coverage missions, aerial data collectors, and communication assistance \cite{10840246}. However, UAVs face significant communication challenges that continue to restrict their performance and broaden their scope of applications.

The broadcast nature of UAV communications, combined with the dominance of line-of-sight (LoS) propagation in UAV-ground channels, makes them more susceptible to jamming attacks, which poses significant security challenges \cite{9322583, 9741304}. Techniques such as UAV cooperative relay communication, trajectory optimization, power control, and other resource allocation strategies can effectively help resist jamming attacks and enhance security performance \cite{9449830}. Owing to the UAV’s controllable trajectory, combating jammers in the spatial domain by maneuvering away from the jamming areas is an attractive option \cite{9454372}. Leveraging UAV's flexibility and freedom of movement to design anti-jamming flight paths is a powerful strategy for safeguarding legitimate communications in UAV-enabled systems \cite{9271902}.
 
By leveraging flexible mobility, anti-jamming trajectories were developed in \cite{9271902, 8555700, 8943423, 8675928} to protect legitimate transmissions in UAV-enabled systems from interference caused by jammers. However, these studies relied on conventional trajectory optimization methods that depend on complete prior knowledge of system parameters. Such methods often suffer from slow convergence, making them impractical for intelligent and autonomous UAVs operating in real-time within dynamic environments. In contrast, AI-based anti-jamming trajectory design using reinforcement learning (RL) offers a more efficient and adaptable solution, as demonstrated in \cite{9849763, 9838325, 10419848}. Although RL methods can quickly generate planning solutions once the network is trained, the training process itself demands a substantial number of samples and multiple training epochs, which require significant time investments. Additionally, many RL methods struggle to adapt to new situations, often necessitating extensive retraining \cite{10446575}. This challenge can hinder real-time prediction and decision-making. Excitingly, active inference emerges as a strong alternative, providing a comprehensive framework for understanding how living organisms interact with their environments \cite{Friston2013-FRIAIA-2}. It emphasizes modeling perception, learning, and decision-making to maximize Bayesian model evidence or minimize free energy.

This work presents an innovative scheme for designing UAV anti-jamming trajectories based on Bayesian active inference, addressing the limitations of existing reinforcement learning (RL) methods. Specifically, we utilize an expert optimizer during training to generate several training examples, which help us develop an efficient UAV trajectory for specific missions, both in the presence and absence of jamming. We construct a world model that encodes the expert's adopted strategy for tackling these training examples and how the expert interacted with the environment to avoid jamming. During testing, the UAV employs autonomous decision-making, enabling it to select an appropriate plan for the current mission, adapt to new situations not encountered during training, and effectively navigate critical challenges such as jamming. A key contribution of our work is the use of an optimizer to derive demonstration results from which we learn the world model. This approach minimizes unnecessary exploration and enhances efficiency, thus leveraging the model's contextual capabilities.

\section{System Model and Problem Formulation}
\begin{figure*}
    \centering
    \includegraphics[width=1.0\linewidth]{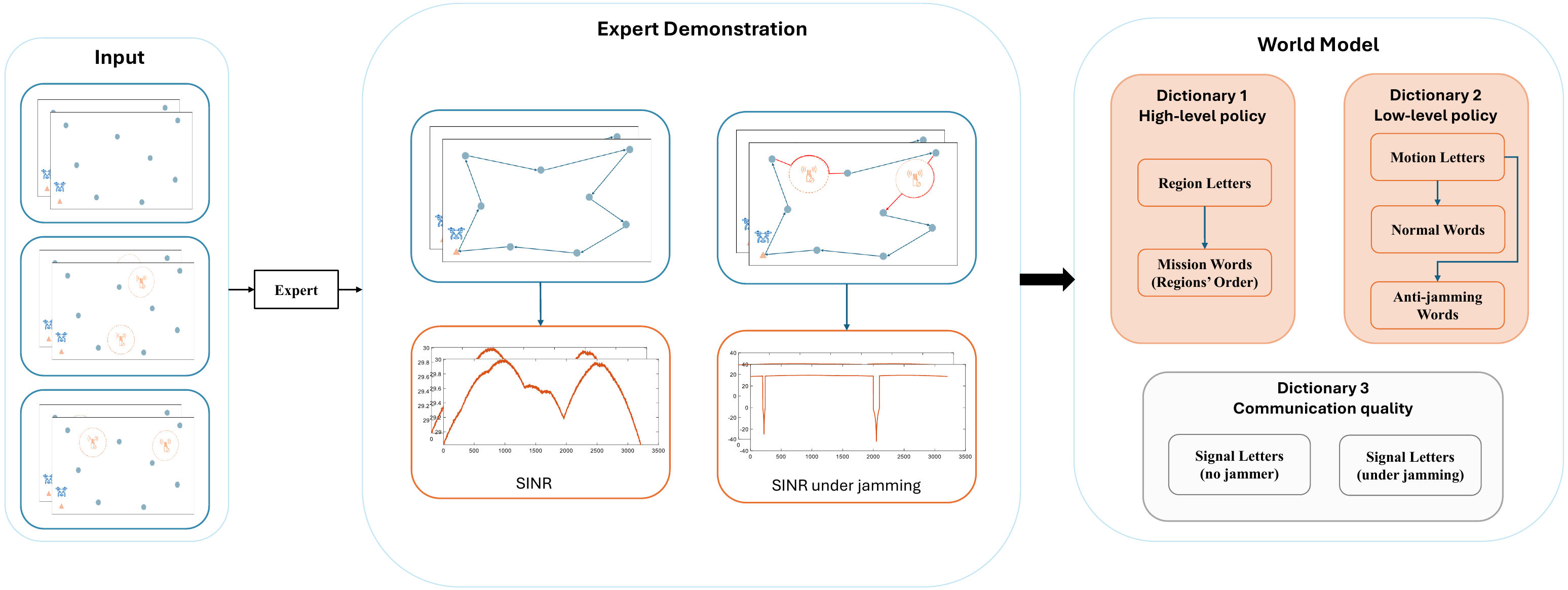}
    \caption{Schematic representing the main steps of the offline process to build the World Model.}
    \label{fig_proposedFrameworkOffline}
\end{figure*}
We consider a UAV-enabled communication system consisting of a single UAV $u$, a control base station (CBS) $b$, a set of ground regions $\mathcal{N} = \{1, 2, \dots, N\}$ that the UAV must visit, and a set of ground-based jammers $\mathcal{J} = \{1, 2, \dots, J\}$. The UAV flies at a fixed altitude $H$ within a square area of side length $D$, where the target regions are randomly located. Communication with the CBS takes place in discrete time slots $t = 1, 2, \dots, T$, and may be disrupted by the jammers.

The wireless channel between the CBS and the UAV is modeled as a ground-to-air link comprising line-of-sight (LoS) and non-line-of-sight (NLoS) components, while small-scale fading is neglected due to its relatively minor influence. The channel path loss is given by:
\begin{equation}
PL(b,u) = 
\begin{cases}
\beta_{\mathrm{LoS}} |d_{bu}|^{-\alpha}, & \text{LoS}, \\
\beta_{\mathrm{NLoS}} |d_{bu}|^{-\alpha}, & \text{NLoS},
\end{cases}
\end{equation}
where $d_{bu} = \sqrt{(x_b - x_u)^2 + (y_b - y_u)^2 + z_u^2}$ is the 3D distance between the CBS and the UAV, and $\alpha$ is the path-loss exponent.

The probability of LoS communication depends on the elevation angle $\theta_b$, modeled as:
\begin{equation}
P_{\mathrm{LoS}} = \frac{1}{1 + \Phi \exp(-\Psi[\theta_b - \Phi])}, \quad
\theta_b = \frac{180}{\pi} \arcsin\left(\frac{z_u}{d_{bu}}\right),
\end{equation}
with typical urban parameters $\Phi = 150$ and $\Psi = 15$.

Jammers are randomly located on the ground and interfere with the UAV's reception when it enters their effective range. The received jamming power from jammer $j$ is given by:
\begin{equation}
I_j = p_j |d_{ju}|^{-\alpha},
\end{equation}
where $p_j$ is the jammer’s transmit power, and $d_{ju} = \sqrt{(x_j - x_u)^2 + (y_j - y_u)^2 + z_u^2}$ is the distance between jammer $j$ and the UAV.

The resulting SINR at the UAV is:
\begin{equation}
\Gamma_u = \frac{p_b(P_{\mathrm{LoS}}\beta_{\mathrm{LoS}} + P_{\mathrm{NLoS}}\beta_{\mathrm{NLoS}})|d_{bu}|^{-\alpha}}{\sum_j I_j + \sigma^2},
\end{equation}
where $p_b$ is the CBS transmit power and $\sigma^2$ is the noise power.

The UAV's mission is modeled as a traveling salesman problem (TSP) where it must visit all target regions while minimizing both the total travel distance and exposure to jamming. The 3D distance between any two regions $n$ and $m$ is $d_{nm} = \sqrt{(x_n - x_m)^2 + (y_n - y_m)^2 + H^2}$.

The optimization problem is formulated as:
\begin{subequations}
\begin{equation} \label{eq_optimizationProblem}
\min_{\{\alpha_{nm}, \beta_j(t)\}} \lambda_1 \sum_{n \neq m} d_{nm} \alpha_{nm} + \lambda_2 \sum_{t=1}^{T} \sum_{j=1}^{J} I_j \beta_j(t) \tag{5}
\end{equation}
\begin{align}
\text{s.t.} \quad & \sum_{m \neq n} \alpha_{nm} = 1, \quad \forall n \in \mathcal{N}, \tag{5a} \label{eq_optimizationProblem_sub1} \\
& \sum_{n \neq m} \alpha_{nm} = 1, \quad \forall m \in \mathcal{N}, \tag{5b} \label{eq_optimizationProblem_sub2} \\
& \beta_j(t) \in \{0, 1\}, \quad \forall j, t, \tag{5c} \label{eq_optimizationProblem_sub3} \\
& \Gamma_u(t) \geq \Gamma_{\min}, \quad \forall t. \tag{5d} \label{eq_optimizationProblem_sub4}
\end{align}
\end{subequations}

In the above formulation, $\alpha_{nm} \in \{0,1\}$ is a binary decision variable indicating whether the UAV travels directly from region $n$ to $m$. Constraints (5a) and (5b) ensure that each region is visited exactly once and that the UAV returns to the starting location, forming a complete tour. 

The variable $\beta_j(t)$ indicates whether the UAV is within the interference range of jammer $j$ at time $t$, modeled as:
\begin{equation}
\beta_j(t) = 
\begin{cases}
1, & \text{if } d_{ju}(t) \leq d_{\text{threshold}}, \\
0, & \text{otherwise}.
\end{cases}
\end{equation}

Constraint (5c) enforces the binary nature of $\beta_j(t)$. Constraint (5d) guarantees communication reliability by requiring the SINR to remain above a predefined threshold $\Gamma_{\min}$ at all times. The objective function balances the trade-off between minimizing travel distance and reducing communication interference, weighted by coefficients $\lambda_1$ and $\lambda_2$.

\section{Proposed Anti-Jamming UAV Trajectory Design Based on Bayesian Active Inference}

\begin{figure*}
    \centering
    \includegraphics[width=1.0\linewidth]{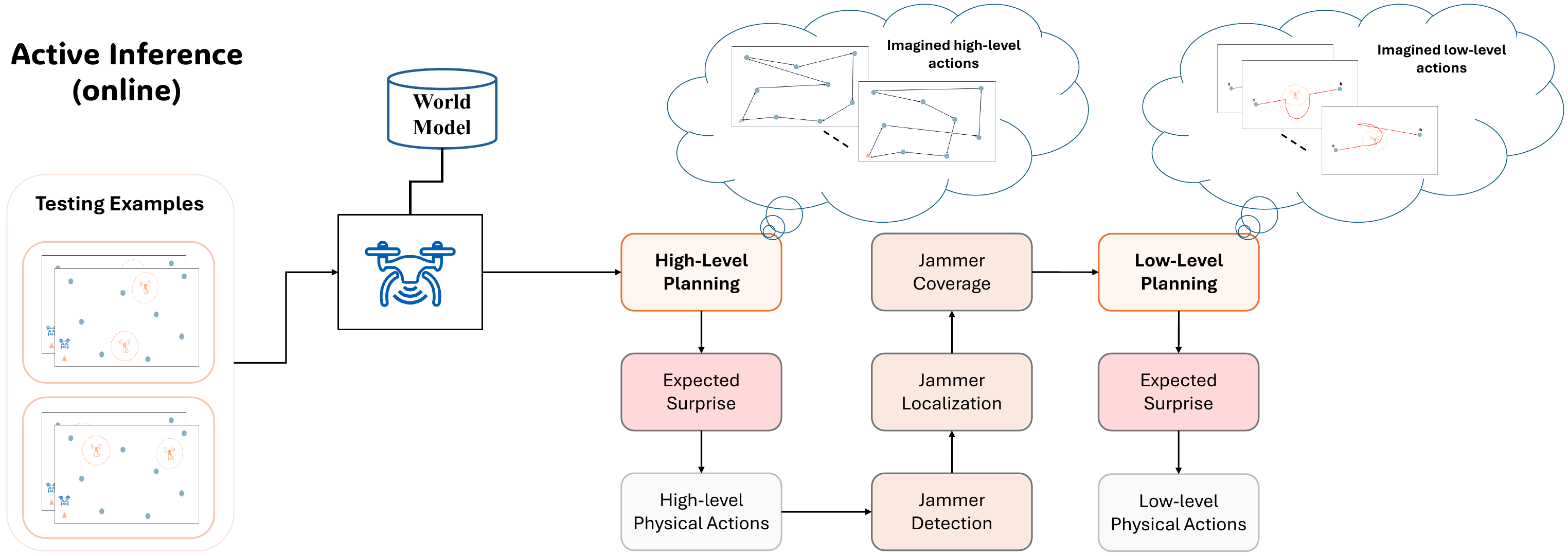}
    \caption{Schematic of the main steps in the Active Inference online process.}
    \label{fig_proposedFrameworkOnline}
\end{figure*}

We propose a Bayesian Active Inference framework that enables UAVs to plan and adapt trajectories robustly under jamming threats. The method comprises three key stages, combining offline learning from expert demonstrations with online probabilistic inference (see Fig.~\ref{fig_proposedFrameworkOffline} and Fig.~\ref{fig_proposedFrameworkOnline}).

\textbf{Expert Demonstration Generation:} First, reference trajectories are computed by solving the optimization problem in \eqref{eq_optimizationProblem} under both normal and jammed conditions. These demonstrations encode near-optimal behaviors for different interference scenarios and provide the data needed to learn robust decision-making patterns.

\textbf{Hierarchical World Model Construction:} Using the demonstrations, the UAV builds a generative world model that reflects how its actions lead to changes in position and signal quality. The model includes: (i) a high-level policy mapping symbolic action sequences over regions; (ii) a low-level planner encoding executable paths via motion primitives; and (iii) an outcome model predicting sensory consequences like SINR changes. This structured model enables the UAV to simulate and evaluate candidate trajectories before execution.

\textbf{Online Inference and Trajectory Adaptation:} During operation, the UAV performs active inference over the learned model. By minimizing expected free energy, it selects high-level plans that balance goal achievement and uncertainty reduction. At the motion level, a Kalman filter–based attractor generates smooth trajectories that follow symbolic guidance while responding to noise. Signal observations update the UAV's beliefs in real time, allowing it to detect interference and adapt its path dynamically.

Overall, this three-stage architecture equips the UAV with the ability to reason over actions and sensory outcomes, resulting in adaptive, communication-aware navigation even in adversarial environments.

\subsection{Expert Demonstration Generation}

The wireless environment is modeled as a graph $\mathcal{G}_n = (\mathcal{V}_n, \mathcal{E}_n)$, where each vertex $\mathrm{v}_m \in \mathcal{V}_n$ represents a ground region at position $p_{\mathrm{v}_m} = [x_{\mathrm{v}_m}, y_{\mathrm{v}_m}]$, and each edge $(n, m) \in \mathcal{E}_n$ denotes a viable transition with associated cost:
\begin{equation}
c_{nm} = 
\begin{cases}
d_{nm}, & \text{if } \mathcal{H}_0 \\
d_{nm} + \sum_{t=1}^{T} \sum_{j=1}^{J} I_j, & \text{if } \mathcal{H}_1
\end{cases}
\end{equation}
Here, $\mathcal{H}_0$ and $\mathcal{H}_1$ correspond to nominal and adversarial (jamming) conditions, respectively. For each of the $K$ missions, an expert planner (e.g., 2-OPT) generates optimal solutions under both hypotheses, resulting in datasets $\mathcal{D}_0 = \{\mathcal{G}_k^{(0)}\}_{k=1}^{K}$ and $\mathcal{D}_1 = \{\mathcal{G}_k^{(1)}\}_{k=1}^{K}$. Each expert demonstration is expressed as a tuple $L_k^{(h)} = (L_k^{(h,\text{high})}, L_k^{(h,\text{low})})$ for $h \in \{0, 1\}$, where the high-level component encodes the region visitation sequence, and the low-level component describes the corresponding motion trajectory. Under $\mathcal{H}_1$, the region sequence is preserved, but the trajectory deviates to circumvent jamming zones.

\textbf{Dictionary 1} captures high-level symbolic structure from the demonstrations in $\mathcal{L}_0 = \{L_k^{(0)}\}_{k=1}^{K}$. Each region is treated as a letter $l_n$, and transitions are defined as directed edges $e(l_n, l_m)$, forming generalized letters $\Tilde{l}_n = [l_n, e(l_n, l_m)]$. The mission is encoded as a word $w_k = \{\Tilde{l}_i\}_{i=1}^{n}$. This structure forms a Generalized Dynamic Bayesian Network (GDBN) that models high-level navigation patterns under nominal conditions.

\textbf{Dictionary 2} represents low-level motion strategies derived from both nominal and jammed scenarios. A normal motion word under $\mathcal{H}_0$ is defined as $w_k^{v,0} = \{l_i^{v,0}\}_{i=1}^{n_0}$, where $l_i^{v,0} = [v_x^i, v_y^i, v_z^i]$ encodes UAV velocity components at each step. Under $\mathcal{H}_1$, anti-jamming motion words $w_k^{v,1} = \{l_i^{v,1}\}_{i=1}^{n_1}$ capture evasive strategies using deflected or curved trajectories. These words are grouped into token sets indexed by jammer characteristics:
\begin{equation}
\mathcal{T}_j = \{w_k^v \mid \text{type } j, r_j, p_j\}, \quad
\mathcal{D}_{\text{anti-jamming}} = \bigcup_{j=1}^{J} \mathcal{T}_j,
\end{equation}
where $p_j$ denotes the jammer’s transmit power and $r_j$ denotes its corresponding coverage radius.
This dictionary enables the learning of modular low-level controllers that adapt to varying interference conditions.

\textbf{Dictionary 3} encodes signal perception during low-level motion. At each step, the UAV records the Signal-to-Interference-plus-Noise Ratio (SINR), which is discretized using an unsupervised clustering algorithm (e.g., Growing Neural Gas) into signal letters $\gamma_i = [\Gamma_i, \Dot{\Gamma}_i]$, where $\Gamma_i$ is the SINR value and $\Dot{\Gamma}_i$ is its temporal derivative. These letters are organized into two sets: $\Upsilon_0 = \{\gamma_i \mid \mathcal{H}_0\}$ for normal conditions and $\Upsilon_1 = \{\gamma_i \mid \mathcal{H}_1\}$ for jamming scenarios. This representation links control actions to communication outcomes, enriching the behavioral model with environment-aware signal feedback.

Together, the three dictionaries define a structured representation of expert behavior at multiple levels of abstraction. \textbf{Dictionary 1} models symbolic region sequences and their transitions; \textbf{Dictionary 2} captures low-level motion control under both nominal and adversarial conditions; and \textbf{Dictionary 3} characterizes the sensory consequences of actions through discretized signal observations. This multi-level decomposition enables separate learning of high-level planning and low-level control, while grounding both in realistic signal feedback experienced by the UAV.

\subsection{Hierarchical World Model Construction}
\begin{figure}
    \begin{minipage}[b]{0.32\linewidth}
     \centering
        \centerline{\includegraphics[width=1.9cm]{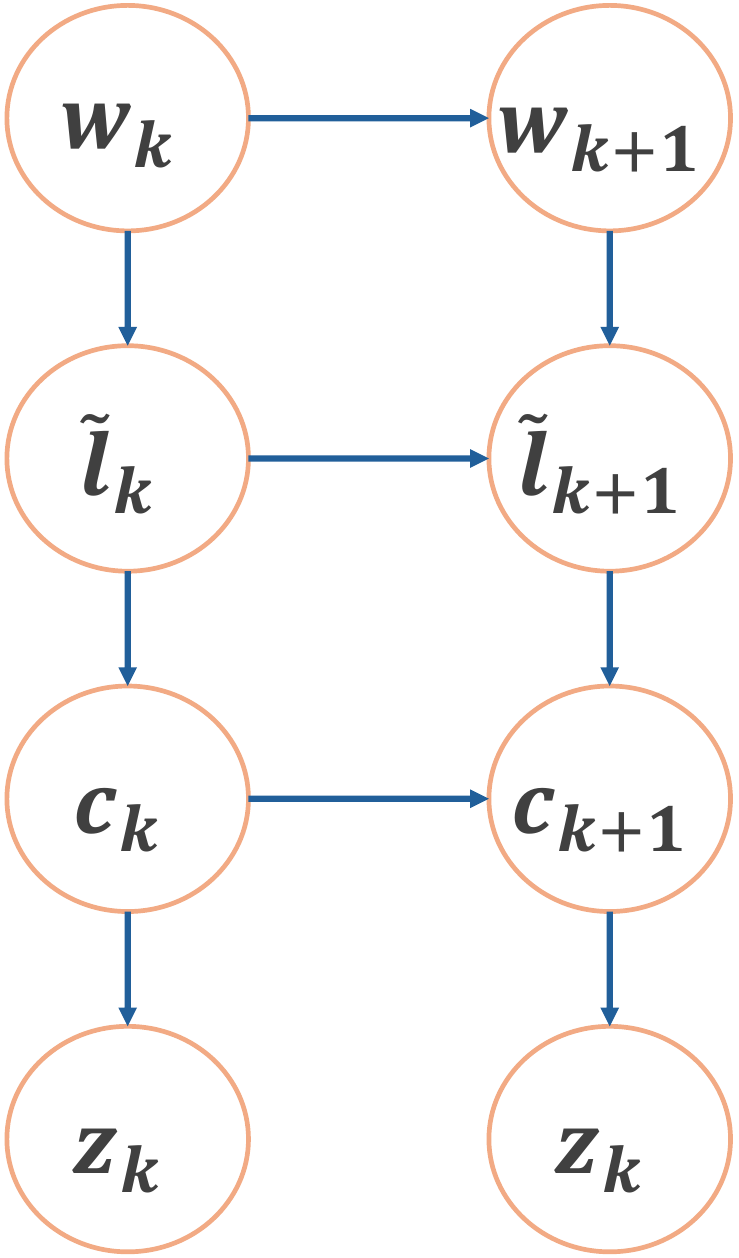}}
        \centerline{\scriptsize (a)}\medskip
    \end{minipage}
    \begin{minipage}[b]{0.32\linewidth}
     \centering
        \centerline{\includegraphics[width=1.9cm]{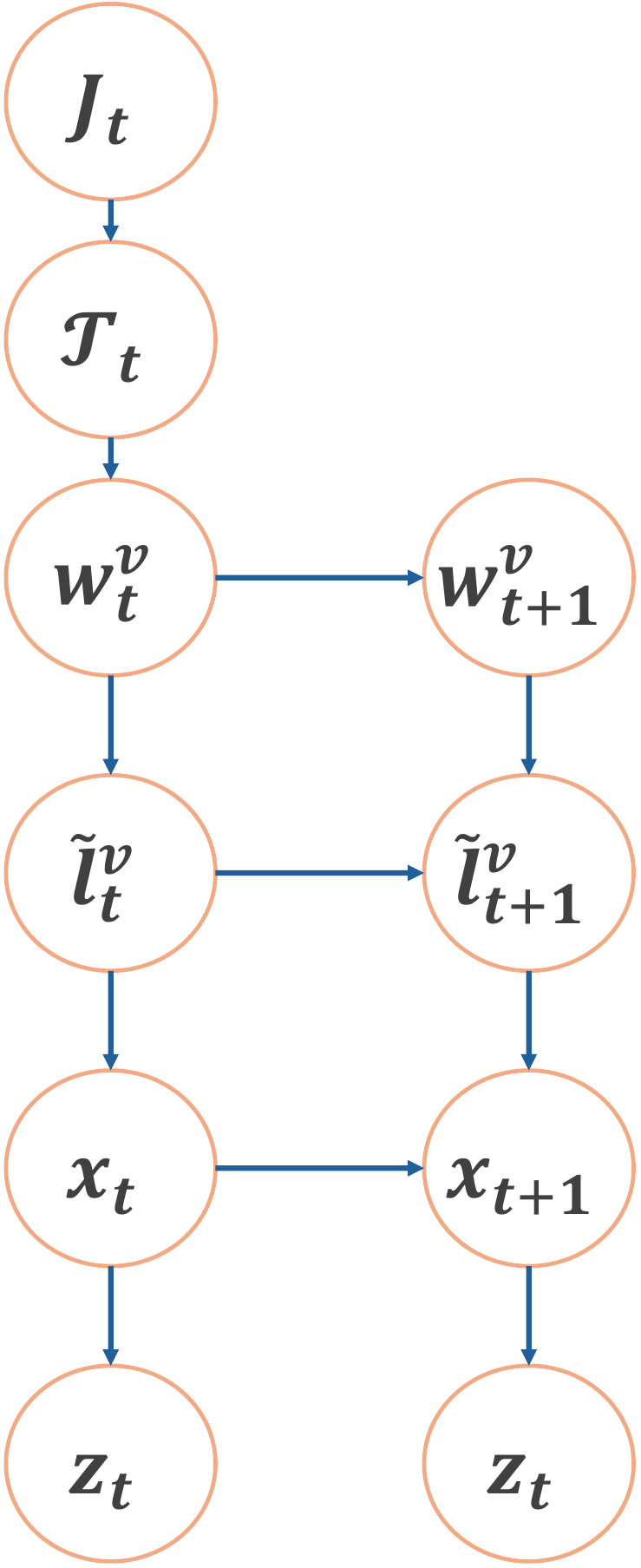}}
        \centerline{\scriptsize (b)}\medskip
    \end{minipage}
    \begin{minipage}[b]{0.32\linewidth}
     \centering
        \centerline{\includegraphics[width=1.9cm]{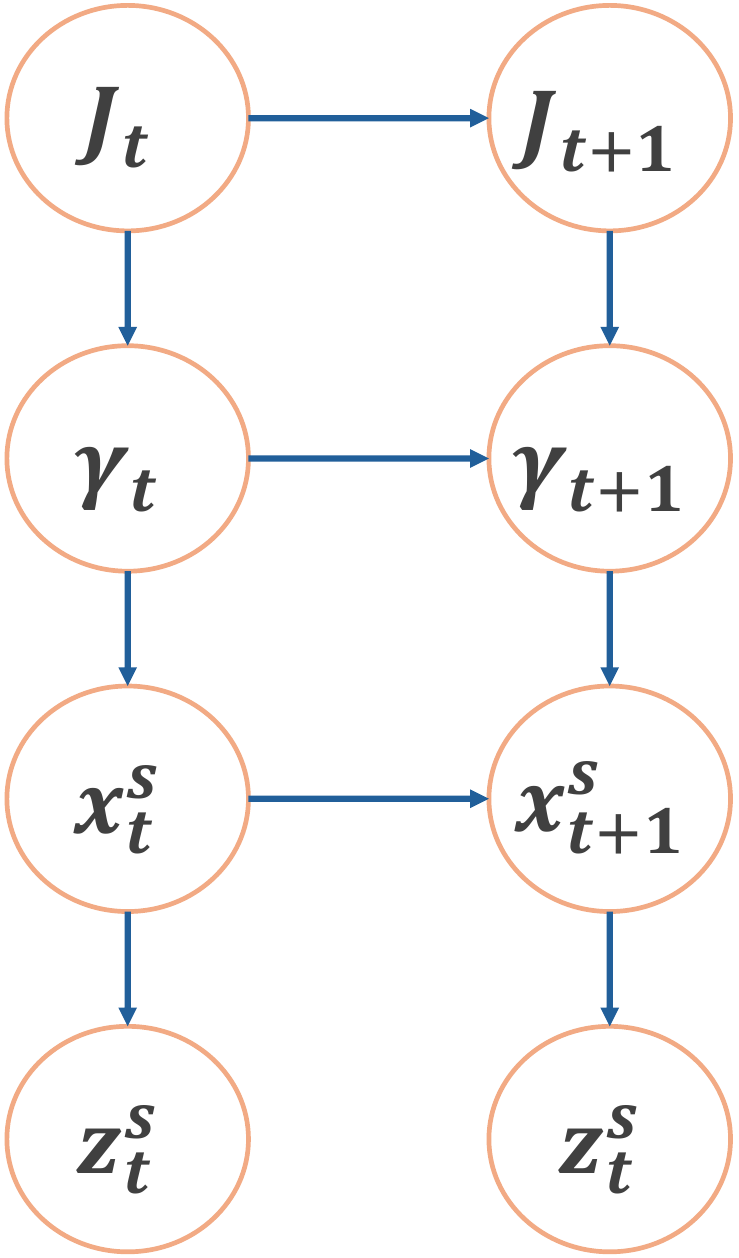}}
        \centerline{\scriptsize (c)}\medskip
    \end{minipage}
    \caption{Dictionaries structured as GDBNs: (a) Dictionary 1, (b) Dictionary 2, (c) Dictionary 3.}
\label{Fig_GDBNs}
\end{figure}
To model expert demonstrations in a structured and probabilistic way, we adopt a hierarchical framework based on Generalized Dynamic Bayesian Networks (GDBNs), as shown in Fig. \ref{Fig_GDBNs}, integrating symbolic planning, motion control, and signal feedback. The model comprises three levels:

\textbf{Dictionary 1} encodes symbolic mission plans as region visitation sequences. Each plan is represented as a word $w_k = \{ \tilde{l}_i \}_{i=1}^{n}$, where $\tilde{l}_i = [l_i, e(l_i, l_{i+1})]$ includes the current region and the transition to the next. The total cost of the mission is:
\begin{equation}
z_k = \sum_{i=1}^{n} c(l_i, l_{i+1})
\end{equation}
The GDBN includes four layers: the symbolic plan $w_t$, transition $\tilde{l}_t$, expected cost $c_t$, and cumulative cost $z_t$. This supports efficient selection of high-level plans with minimal expected cost.

\textbf{Dictionary 2} captures low-level motion policies as velocity words $w_k^v = \{ l_i^v \}_{i=1}^{n'}$, with $l_i^v = [v_x^i, v_y^i, v_z^i]$. These encode either straight or evasive trajectories depending on the presence of jammers. Words are grouped into tokens $\mathcal{T}_j$ defined by jammer characteristics $(r_j, p_j)$:
\begin{equation}
\mathcal{T}_j = \{w_k^v \mid \text{jammer type } j\}
\end{equation}
A latent variable $j_i$ selects the active token:
\begin{equation}
\mathcal{T}_i = 
\begin{cases}
\mathcal{T}_j, & \text{if } j_i = j \\
\mathcal{T}_0, & \text{otherwise}
\end{cases}
\end{equation}
A sampled word generates velocity commands, and UAV motion is modeled as:
\begin{equation}
\dot{x}_t = -K_v (\dot{x}_t - l_t^v) + \epsilon_t, \quad \epsilon_t \sim \mathcal{N}(0, \Sigma)
\end{equation}
In discrete time, this becomes:
\begin{equation} \label{eq_dynamic_model_attractor}
x_{t+1} = x_t + u_t + \epsilon_t, \quad u_t = \alpha l_t^v \Delta t, \quad \alpha = \frac{K_v}{1 + K_v}
\end{equation}
where $u_t$ with $\alpha \in (0,1)$ modulating the strength of attraction, and $\epsilon_t$ captures stochastic variability.
This formulation treats symbolic velocities as attractors, enabling robust trajectory execution without predefined waypoints.

\textbf{Dictionary 3} models signal feedback by clustering SINR observations into symbolic signal letters $\gamma_t = [\Gamma_t, \Dot{\Gamma}_t]$, representing SINR and its temporal change. Letters are categorized as nominal or jammed, based on the jammer indicator $J_t \in \{0, 1\}$. The GDBN links $J_t$ to $\gamma_t$, a latent SINR state $x_t$, and the observed signal $z_t$. This allows the UAV to interpret noisy measurements and adapt behavior based on communication quality.

Together, the three dictionaries form a hierarchical model that unifies symbolic planning, low-level control, and perception. This structure supports adaptive, interpretable, and robust UAV decision-making under both nominal and adversarial conditions.

\subsection{Bayesian Active Inference for Online Trajectory Adaptation}

\paragraph{Online High-Level Planning}

During deployment, the UAV must plan symbolic routes over dynamic environments by visiting mission-critical regions. This high-level task is modeled using the GDBN structure from Dictionary 1, which encodes expert demonstrations as symbolic region-visitation sequences.

Let the current mission be defined by an unordered set of target regions:
\[
w_{t}^{\text{testing}} = \{l_1, l_2, \dots, l_n\}
\]
where each $l_i$ denotes a region. This set may evolve as targets appear or vanish. The UAV aims to infer a plausible ordered sequence $\hat{w}$ that meets the mission goal while aligning with prior expert knowledge.

To achieve this, the UAV evaluates a candidate set $\mathcal{W}_t = \{\hat{w}_1, \dots, \hat{w}_K\}$, consisting of feasible permutations of $w_t$, constrained by structural priors (e.g., preserved subsequences). Each candidate $\hat{w}_k$ is treated as a hypothesis and passed through the generative model to predict the expected latent transition and cumulative cost distributions:
\begin{equation}
\pi(\tilde{l}_{t}) = P(\tilde{l}_{t}|\tilde{l}_{t-1}, \hat{w}_{k}), \quad \pi(c_{t}) = P(c_{t}|c_{t-1}, \tilde{l}_{t}) \cdot \pi(\tilde{l}_{t}).
\end{equation}

Each hypothesis is then compared to a trusted reference plan $\pi(c_{\text{ref}})$ from Dictionary 1. This comparison uses KL divergence to compute an abnormality score:
\begin{equation}
A(\hat{w}_k) = D_{\text{KL}} \left( \pi(c_t) \,\|\, \lambda(c_t) \right),
\end{equation}
where $\lambda(c_t) = P(z_t|c_t)$ is the diagnostic distribution induced by observations. A higher $A(\hat{w}_k)$ implies a greater deviation from expert behavior, indicating greater epistemic uncertainty.

Following the Active Inference principle, the UAV selects the candidate that minimizes this abnormality:
\begin{equation}
\hat{w}^{*} = \arg \min_{\hat{w}_k \in \mathcal{W}_t} A(\hat{w}_k).
\end{equation}
This ensures that the chosen plan is both contextually grounded and statistically aligned with successful prior knowledge, enabling robust symbolic planning under uncertainty.

\paragraph{Online Low-Level Planning and Signal-Driven Jamming Response}

Once $\hat{w}^*$ is selected, the UAV enters the low-level planning phase. For each segment between consecutive regions, e.g., from $l_1^*$ to $l_2^*$, a velocity word $w_m^v$ from Dictionary 2 is chosen, composed of letters $\{l_i^v\}$ encoding discrete velocity vectors.

The UAV incorporates these targets within a Kalman filter, using each $l_i^v$ as an attractor in the discrete-time dynamics of Eq.~\eqref{eq_dynamic_model_attractor}, which supports smooth and uncertainty-aware motion execution.

Simultaneously, the UAV predicts expected SINR values $\hat{\gamma}_t$ via Dictionary 3. At each time step, the observed signal $z_t$ is compared with the prediction using:
\begin{equation}
A_t^{\text{sig}} = D_{\text{KL}}\left( \pi(\hat{\gamma}_t) \,\Vert\, \lambda(\hat{\gamma}_t) \right).
\end{equation}
Large values of $A_t^{\text{sig}}$ suggest potential jamming. The UAV then initiates inference over a set of candidate jammer particles $\{\theta_j\}$, each representing a hypothetical jammer location $q_j = [x_j, y_j]$ and radius $r_j$.

For each $\theta_j$, a simulated SINR trajectory $\hat{\gamma}_t^{(j)}$ is generated, and its cumulative abnormality is evaluated:
\begin{equation}
A^{\text{jam}}(\theta_j) = \sum_t D_{\text{KL}}\left( \pi(\hat{\gamma}_t^{(j)}) \,\Vert\, \lambda(\hat{\gamma}_t) \right).
\end{equation}
The most likely jammer is inferred as:
\begin{equation}
\theta^* = \arg\min_{\theta_j} A^{\text{jam}}(\theta_j).
\end{equation}

The corresponding motion token $\mathcal{T}_{\theta^*}$ from Dictionary 2 is selected, containing trajectories optimized for the inferred jammer configuration. The UAV then chooses the trajectory with the lowest expected signal abnormality:
\begin{equation}
w_{\text{anti}}^* = \arg\min_{w_m^v \in \mathcal{T}_{\theta^*}} \sum_{t} A_t^{\text{sig}}(w_m^v).
\end{equation}

This allows the UAV to escape jamming regions while preserving communication reliability. The entire process—combining hierarchical planning, probabilistic signal reasoning, and real-time inference—enables intelligent adaptation to unforeseen wireless threats during mission execution.

\section{Results and Discussion}

This section presents a comprehensive evaluation of the proposed hierarchical planning framework based on Bayesian Active Inference. The objective is to assess its ability to generate efficient and robust UAV trajectories in environments with and without communication interference.

\subsection{System Setup}

The UAV operates in a $1000 \times 1000~\mathrm{m}^2$ 2D environment populated with randomly placed mission-critical nodes that mimic real-world targets such as inspection points or communication hotspots. The UAV maintains a fixed altitude of $200~\mathrm{m}$ and travels at constant speed, with its trajectory discretized at $1~\mathrm{m}$ intervals to ensure precise control.

To model interference, jammers are randomly positioned across the environment. Each emits within a $50~\mathrm{m}$ radius and includes up to $30~\mathrm{m}$ of positional jitter to simulate deployment uncertainty. The number and placement of jammers vary across runs.

Communication occurs over a continuous downlink from a control base station transmitting at $1~\mathrm{W}$ (30 dBm). The wireless channel incorporates probabilistic path loss with LOS probability $0.9$ and NLOS probability $0.1$, path loss exponent $\eta = 2$, and intercepts $\beta_{\text{LOS}} = 1~\mathrm{dB}$ and $\beta_{\text{NLOS}} = 20~\mathrm{dB}$. The UAV continuously estimates SINR, which informs both its high-level planning and low-level control decisions.

Two datasets are used for training: \textbf{Dataset 1} includes trajectories under nominal conditions, and \textbf{Dataset 2} includes examples with jammers. Expert-generated demonstrations are used to construct the three-level dictionary model described earlier, enabling the UAV to learn both symbolic planning strategies and motion-level policies under different communication conditions.

\subsection{Online Trajectory Adaptation}

\begin{figure}
    \begin{minipage}[b]{0.48\linewidth}
     \centering
        \includegraphics[width=4.2cm]{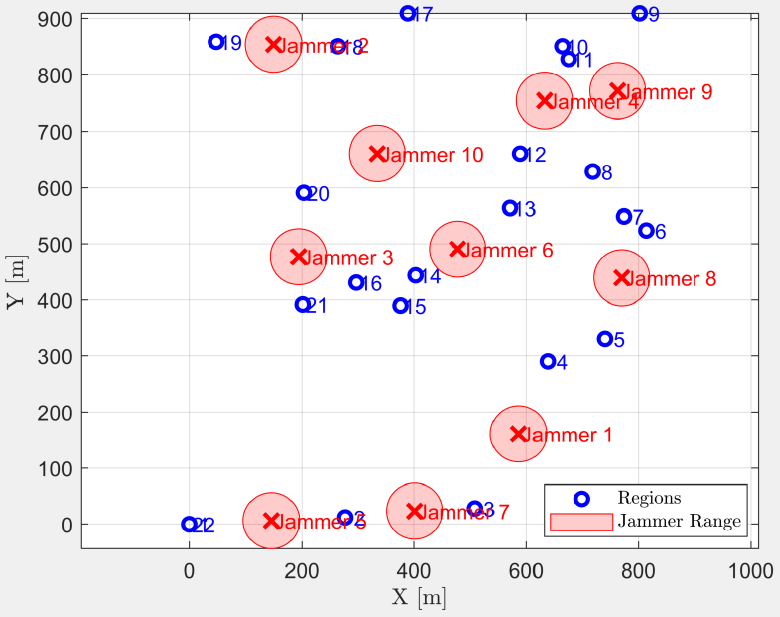}
        \centerline{\scriptsize (a) Testing Example}
    \end{minipage}
    \begin{minipage}[b]{0.48\linewidth}
     \centering
        \includegraphics[width=4.2cm]{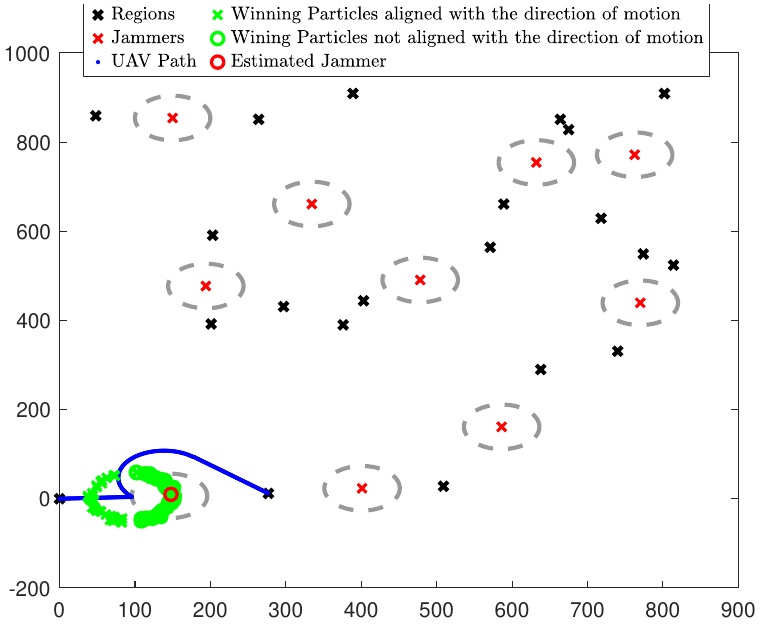}
        \centerline{\scriptsize (b) Event 1}
    \end{minipage}
    \\[4mm]
    \begin{minipage}[b]{0.48\linewidth}
     \centering
        \includegraphics[width=4.2cm]{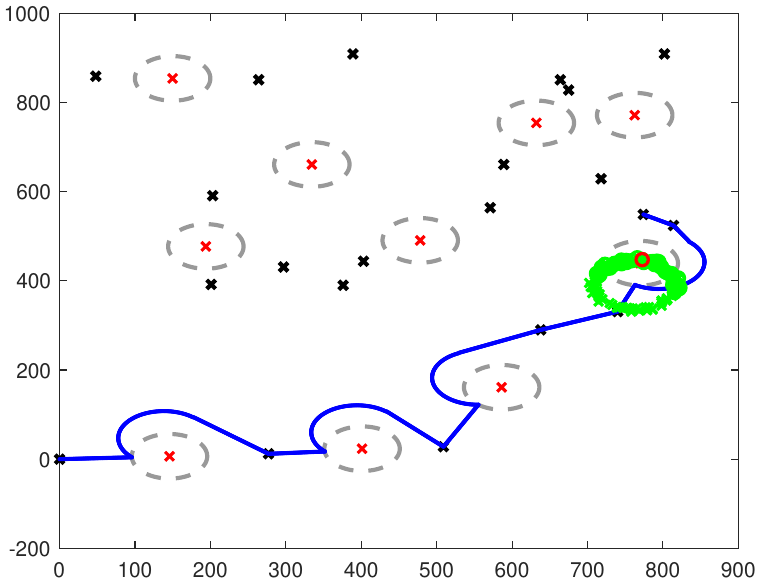}
        \centerline{\scriptsize (c) Event 6}
    \end{minipage}
    \begin{minipage}[b]{0.48\linewidth}
     \centering
        \includegraphics[width=4.2cm]{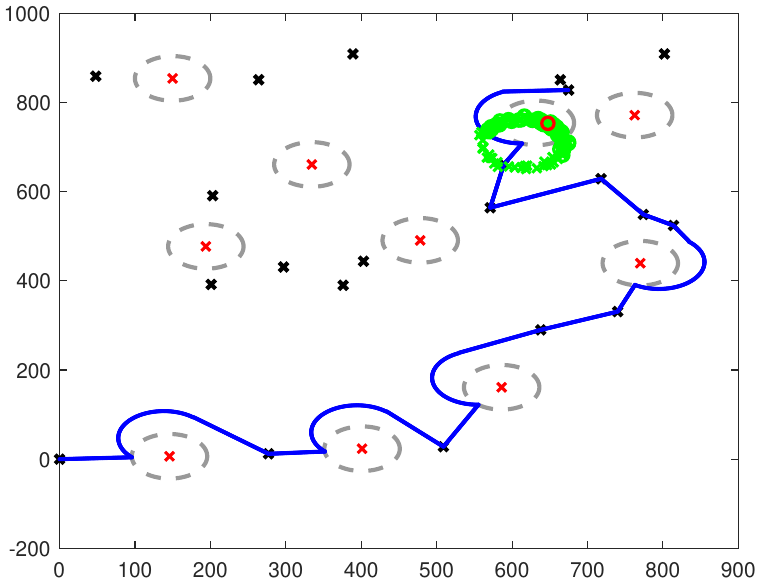}
        \centerline{\scriptsize (d) Event 7}
    \end{minipage}
    \\[4mm]
    \begin{minipage}[b]{0.48\linewidth}
     \centering
        \includegraphics[width=4.2cm]{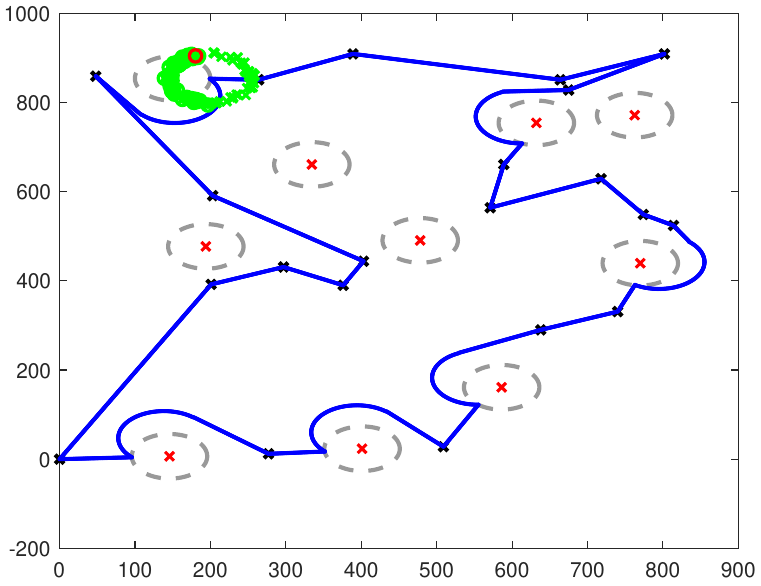}
        \centerline{\scriptsize (e) Event 9}
    \end{minipage}
    \begin{minipage}[b]{0.48\linewidth}
     \centering
        \includegraphics[width=4.2cm]{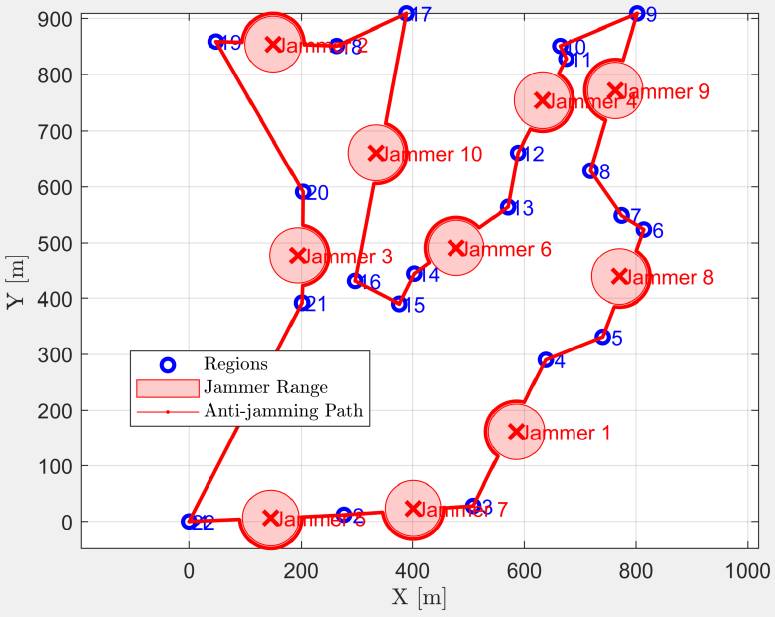}
        \centerline{\scriptsize (f) Expert Anti-jamming trajectory}
    \end{minipage}
    \caption{Online trajectory adaptation under jamming using Bayesian Active Inference.}
    \label{Fig_testingExample_online}
\end{figure}

Fig.~\ref{Fig_testingExample_online} illustrates an example of the UAV performing real-time trajectory adaptation. In Fig.~\ref{Fig_testingExample_online}-(a), the UAV is tasked with visiting a set of blue-marked target regions, unaware of the true locations of red ‘X’-marked jammers. As the UAV moves, it continuously updates its belief about the environment based on observed SINR.

Figs.~\ref{Fig_testingExample_online}-(b) through (e) show how particles are used to infer likely jammer positions. Green crosses indicate particle hypotheses consistent with observed signal degradation, while the red-bordered circle marks the most likely jammer location inferred at that step. The UAV uses this information to select anti-jamming motion policies from Dictionary 2 and re-plan its path accordingly.

In Fig.~\ref{Fig_testingExample_online}-(f), the offline expert trajectory is shown for comparison. Despite not knowing the true jammer positions, the UAV's inferred trajectory closely matches the expert solution, demonstrating its ability to generalize learned behavior to unseen adversarial configurations.

\subsection{Performance Evaluation}

\begin{figure}
\begin{center}
    \begin{minipage}[b]{0.75\linewidth}
     \centering
        \includegraphics[width=5.9cm]{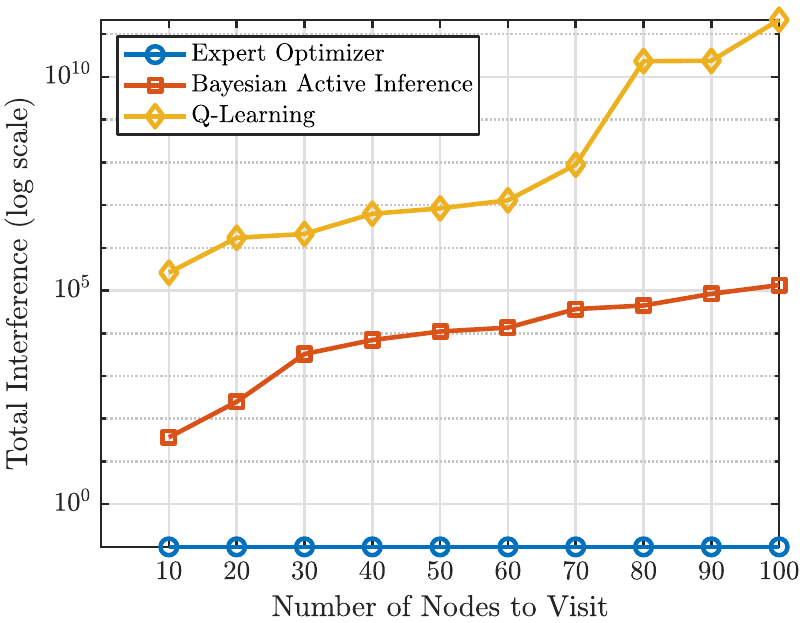}
        \centerline{\scriptsize (a) Total Interference}
    \end{minipage}
    \\[2mm]
    \begin{minipage}[b]{0.65\linewidth}
     \centering
        \includegraphics[width=5.7cm]{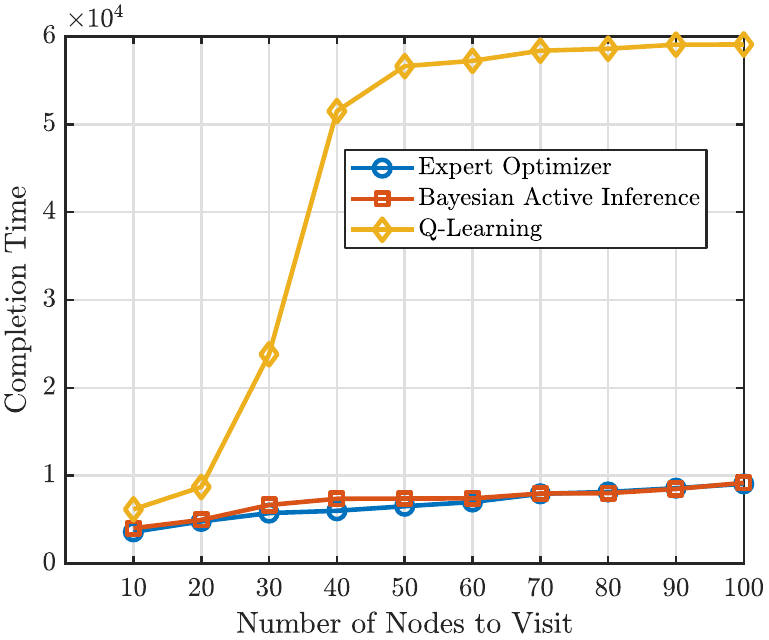}
        \centerline{\scriptsize (b) Completion Time}
    \end{minipage}
\end{center}
\caption{Performance vs. expert and Q-learning across region counts.}
    \label{Fig_performance_comparison}
\end{figure}

Fig.~\ref{Fig_performance_comparison} compares the performance of our method to an expert optimizer and a Q-learning baseline. In Fig.~\ref{Fig_performance_comparison}-(a), we observe that the expert maintains near-zero interference thanks to full knowledge of jammer positions. Q-learning performs poorly as the number of targets increases, reflecting its reactive nature and lack of environmental modeling. In contrast, our framework maintains low interference levels by anticipating jamming zones through probabilistic inference.

Fig.~\ref{Fig_performance_comparison}-(b) shows that the expert achieves the shortest completion times, but Bayesian Active Inference remains close. While Q-learning becomes inefficient as complexity increases, our method scales better by leveraging high-level symbolic planning and low-level adaptation.

\begin{figure}
\begin{center}
    \begin{minipage}[b]{0.65\linewidth}
     \centering
        \includegraphics[width=5.7cm]{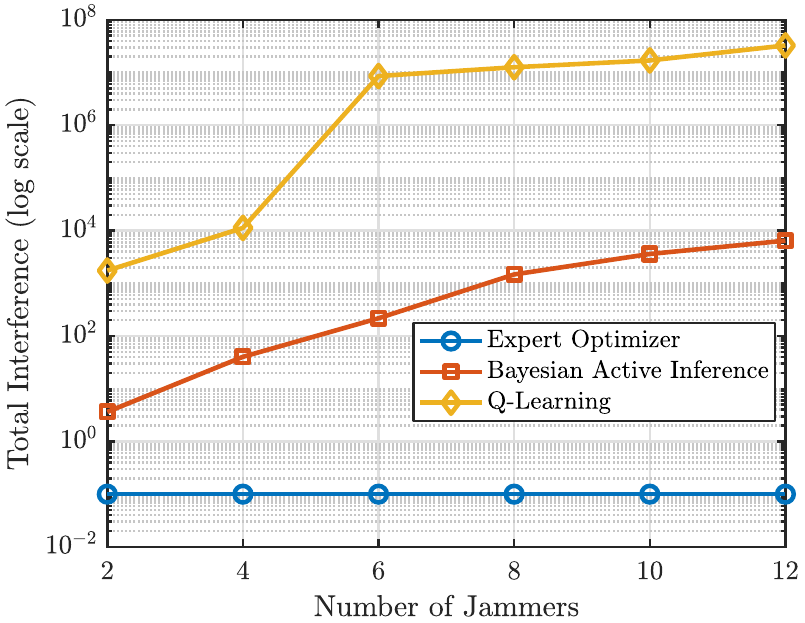}
        \centerline{\scriptsize (a) Total Interference}
    \end{minipage}
    \\[3mm]
    \begin{minipage}[b]{0.65\linewidth}
     \centering
        \includegraphics[width=5.7cm]{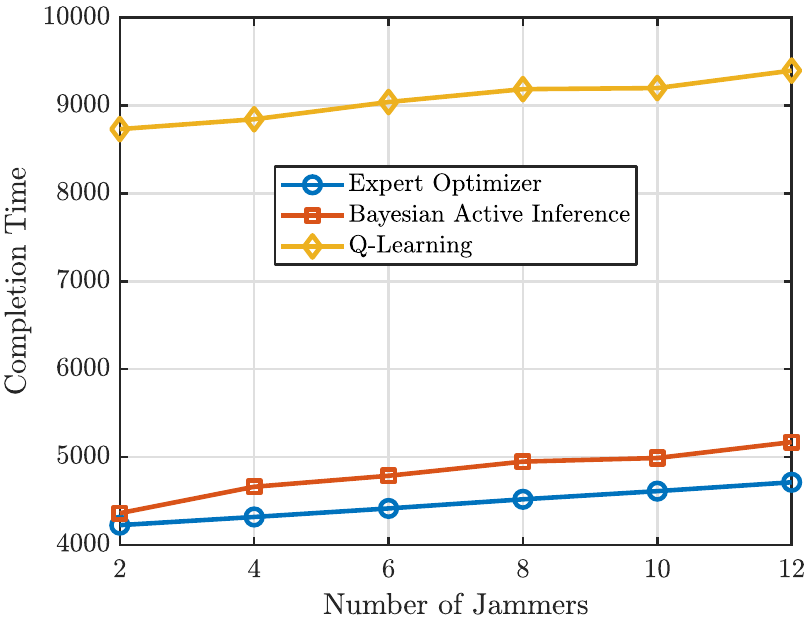}
        \centerline{\scriptsize (b) Completion Time}
    \end{minipage}
\end{center}
\caption{Impact of jammer density on total interference and mission duration.}
    \label{Fig_jammer_scaling}
\end{figure}

Fig.~\ref{Fig_jammer_scaling} evaluates the impact of increasing jammer density. As shown in (a), interference grows significantly for Q-learning, which lacks adaptability. Our method maintains lower interference even in dense environments due to online inference. Fig.~\ref{Fig_jammer_scaling}-(b) shows that mission duration increases modestly for both the expert and our method, while Q-learning struggles to complete tasks under severe interference. These results confirm the robustness and scalability of the proposed approach.

\begin{figure}
\begin{center}
    \begin{minipage}[b]{0.65\linewidth}
     \centering
        \includegraphics[width=5.5cm]{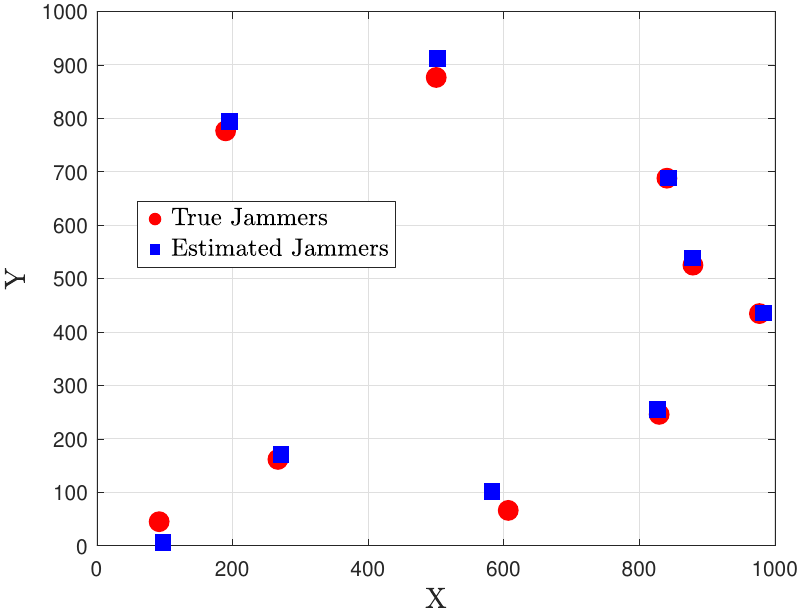}
        \centerline{\scriptsize (a) Estimated Jammers}
    \end{minipage}
    \\[3mm]
    \begin{minipage}[b]{0.65\linewidth}
     \centering
        \includegraphics[width=5.5cm]{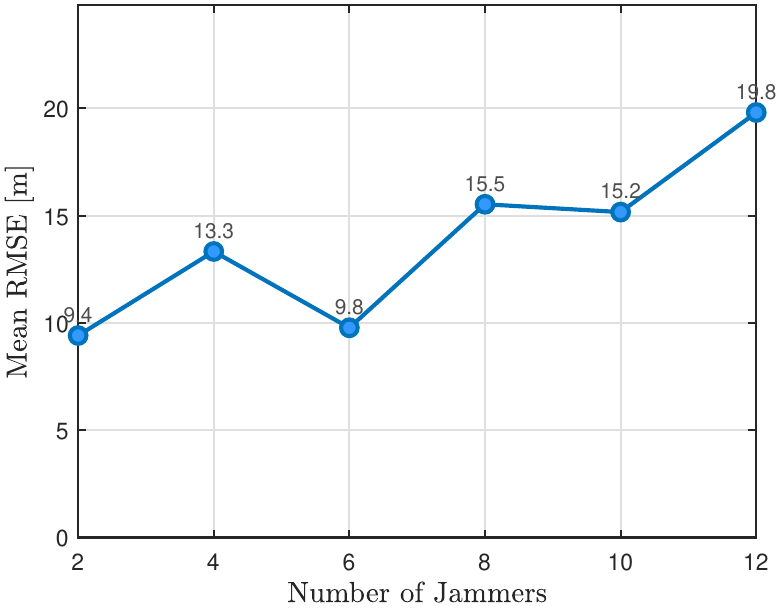}
        \centerline{\scriptsize (b) RMSE}
    \end{minipage}
\end{center}
\caption{Jammer localization accuracy under increasing jammer counts.}
    \label{Fig_estimated_jammers}
\end{figure}

Finally, Fig.~\ref{Fig_estimated_jammers} reports jammer localization accuracy. In Fig.~\ref{Fig_estimated_jammers}-(a), estimated locations (blue squares) are close to true jammer positions (red circles). Fig.~\ref{Fig_estimated_jammers}-(b) shows the RMSE across varying jammer counts, with errors remaining below $20~\mathrm{m}$ in all cases. This confirms that the framework not only plans effectively but also maintains accurate situational awareness in uncertain environments.

Overall, these results highlight the strengths of Bayesian Active Inference: principled integration of planning and perception, online adaptation, and robust generalization under communication threats—all while requiring only limited prior knowledge of environmental dynamics.

\section{Conclusion}
We introduced a Bayesian Active Inference framework for resilient UAV trajectory planning in the presence of jamming threats. By incorporating expert knowledge into a structured generative model, the UAV is able to reason over symbolic mission objectives and adapt low-level motion based on real-time signal observations. The method exhibits strong performance across varying mission scales and jammer densities, achieving low interference, accurate jammer localization, and efficient path planning. These results underscore the potential of hierarchical probabilistic inference for secure and adaptive UAV operations in contested wireless environments. In future work, we aim to extend the framework to additional wireless channel models to further evaluate its robustness.

\section*{Acknowledgment} 
This work was partially supported by the European Union under the Italian National Recovery and Resilience Plan (PNRR) of NextGenerationEU partnership on "Telecommunications of the Future" (PE00000001 - program "RESTART"), CUP E63C22002040007 - D.D. n.1549 of 11/10/2022, and in part by the Ministry of University and Research (MUR), National Recovery and Resilience Plan (NRRP), Mission 4, Component 2, Investment 1.5, project ”RAISE - Robotics and Al for Socio-economic Empowerment” (ECS00000035).

\vspace{12pt}


\bibliographystyle{IEEEtran}
\bibliography{references}

\end{document}